\documentclass[10pt,twocolumn,letterpaper]{article}

\usepackage[pagenumbers]{cvpr} 

\usepackage[symbol]{footmisc}
\usepackage{graphicx}
\usepackage{amsmath}
\usepackage{amssymb}
\usepackage{booktabs}
\usepackage{subcaption}
\usepackage{multirow}
\usepackage{bigstrut}
\usepackage{bbding}
\usepackage{pifont}
\usepackage{enumitem}
\usepackage{float}
\usepackage[accsupp]{axessibility} 
\usepackage{amsfonts,amssymb} 
\usepackage[pagebackref,breaklinks,colorlinks]{hyperref}
\newcommand\graycross{\textcolor[rgb]{ .502,  .502,  .502}{\ding{55}}}
\usepackage[capitalize]{cleveref}
\crefname{section}{Sec.}{Secs.}
\Crefname{section}{Section}{Sections}
\Crefname{table}{Table}{Tables}
\crefname{table}{Tab.}{Tabs.}

\def\cvprPaperID{4935} 
\def\confName{CVPR}
\def\confYear{2023}

\begin{document}

\title{Weakly Supervised Video Representation Learning \\with Unaligned Text for Sequential Videos}


\author{ Sixun Dong$^1$\footnote[1]{} , Huazhang Hu$^1$\footnote[1]{} , Dongze Lian$^{1,2}$,  Weixin Luo$^3$, Yicheng Qian$^1$, Shenghua Gao$^{1,4,5}$\footnote[2]{} \\
$^1$ShanghaiTech University \qquad $^2$National University of Singapore \qquad$^3$Meituan\\
$^4$Shanghai Engineering Research Center of Intelligent Vision and Imaging\\
$^5$Shanghai Engineering Research Center of Energy Efficient and Custom AI IC\\
{\tt\small   \{dongsx, huhzh, liandz, luowx, qianyc, gaoshh\}@shanghaitech.edu.cn}\\
}
\maketitle
\footnotetext[1]{Equal Contribution.} 
\footnotetext[2]{Corresponding Author.}
\begin{abstract}
Sequential video understanding, as an emerging video understanding task, has driven lots of researchers' attention because of its goal-oriented nature. This paper studies weakly supervised sequential video understanding where the accurate time-stamp level text-video alignment is not provided. We solve this task by borrowing ideas from CLIP. Specifically, we use a transformer to aggregate frame-level features for video representation and use a pre-trained text encoder to encode the texts corresponding to each action and the whole video, respectively. To model the correspondence between text and video, we propose a multiple granularity loss, where the video-paragraph contrastive loss enforces matching between the whole video and the complete script, and a fine-grained frame-sentence contrastive loss enforces the matching between each action and its description. As the frame-sentence correspondence is not available, we propose to use the fact that video actions happen sequentially in the temporal domain to generate pseudo frame-sentence correspondence and supervise the network training with the pseudo labels. Extensive experiments on video sequence verification and text-to-video matching show that our method outperforms baselines by a large margin, which validates the effectiveness of our proposed approach. Code is available at \url{https://github.com/svip-lab/WeakSVR}.



\end{abstract}

\section{Introduction}\label{sec:intro}
\begin{figure*}
\centering
\centerline{\includegraphics[width=\textwidth]{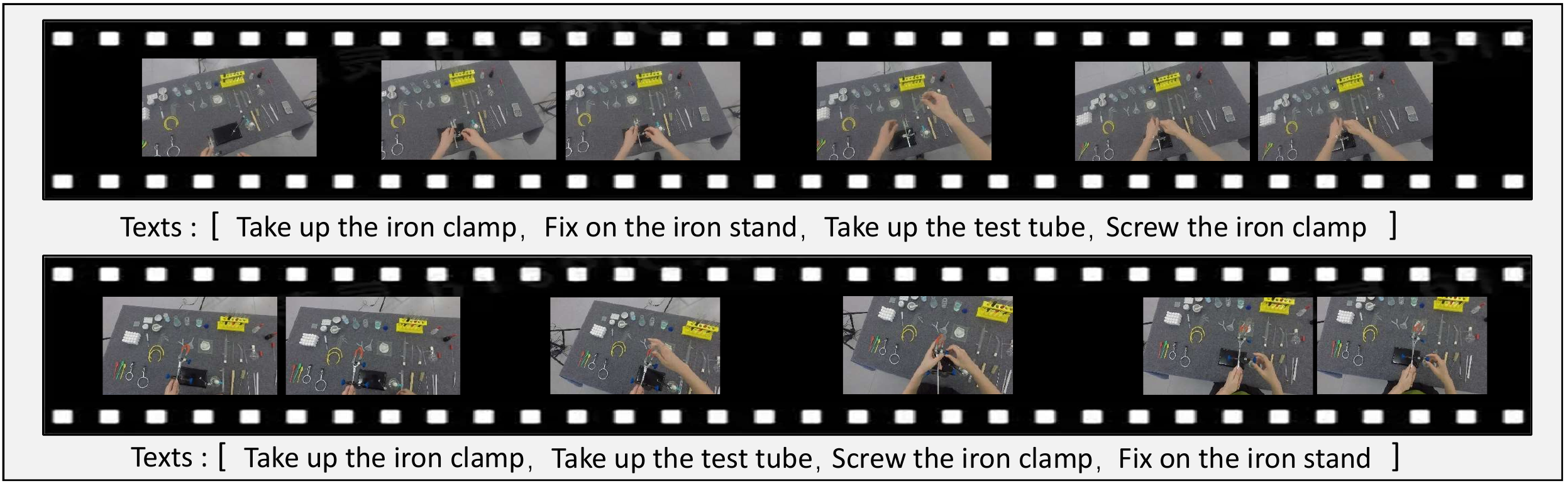}}
\caption{
\textbf{Sequential Video.}
The samples come from CSV dataset. They describe two types of step schedule to accomplish the task of ”fix the test tube on the iron stand with iron clamp”. The upper process the step "fix on the iron stand " before the steps "take up the test tube" and "screw the iron clamp". Diversely, the lower make the steps of "take up the test tube" and "screw the iron clamp" before the step "fix on the iron stand ". It can be seen that the order, time span and temporal location of sub-actions to accomplish the task are apparently different. 
}
\label{fig:sequence_video}
\vspace{-1.0em}
\end{figure*}

A strong artificial intelligence (AI) system is expected to be able to learn knowledge from the open world in an embodied manner such that amounts of goal-oriented tasks are designed for reinforcement learning in the environment. In the area of video understanding, a great deal of pioneering work in video classification \cite{classification}, action localization \cite{localization}, and action segmentation \cite{segmentation} has been explored, laying the foundation for video understanding. Beyond these typical video understanding tasks, sequential videos (such as \cref{fig:sequence_video}) that usually describe how to perform a task in a certain sequence of procedures can be regarded as a goal-oriented task. Solving this task is extremely promising for guiding intelligence to learn a task like humans. It makes performing sequential video representations a potentially critical part of the road to strong AI.

Some efforts have been made for video representation learning for sequential videos. \eg, \cite{Timestamp,Xie2022Alignment} learns a video representation in an instructive video. However, these methods rely heavily on the annotations of temporal boundaries, \, i.e., the timestamps of sequential actions, which are usually difficult to be obtained due to the time-consuming human labeling in practice. A common but often overlooked scenario is that sequential videos usually occur accompanied with audio or text narrations, which show consistent steps with explanations. The rich text information describes the corresponding procedure in detail as shown in \cref{fig:sequence_video}, but they are usually not aligned with videos. Therefore, a question arises, \, i.e., whether it is possible to directly learn the video representation with unaligned text and video in a weakly supervised manner.

With the popularity of visual-language tasks, multi-modal learning has attracted growing attention and has been explored in a variety of areas, \, e.g., image classification \cite{kinetics, C3D}, object detection \cite{CAT-Det, detection}, and video understanding \cite{coca}. One of the most representative works is CLIP\cite{CLIP}. It has shown the potential of learning a powerful semantic representation from natural language supervision with a contrastive learning loss and the strong zero-shot generalization on the downstream tasks, such as text-video retrieval\cite{sun2022long, VLM}, action segmentation \cite{actbert}, multiple-choice videoQA \cite{bridgeformer, VLM} and action step localization\cite{LocalVTP}. VideoCLIP\cite{videoclip} presents a contrastive learning approach to pre-train a unified model with video-text pairs, and \cite{Timestamp} proposes a unified fully and timestamp-supervised framework for multi-model action segmentation. This provides us with an alternative for weakly supervised video representation learning. However, all these previous works are equipped with aligned texts and video frames \cite{Timestamp}, which is not existent in our weakly supervised setting. Thus, it is intractable to directly adapt the existing multi-modal video representation models to our task.  

To overcome the unalignment issue between text and video and learn a satisfactory video representation, we propose a weakly supervised video representation learning pipeline and introduce a multiple granularity contrastive loss to constrain the model, which takes full account of the pseudo temporal alignment between frames and sentences. To be specific, we first extract video and text features from a CLIP-based vision-language model, and a global contrastive loss is designed to constrain the complete video-paragraph alignment. It constrains that a video will be closer to the sequence of the texts describing it while far away from the rest of the texts, and vice versa. Secondly, we introduce a fine-grained contrastive learning loss, which encourages the frame sequences of representations to be more similar to the neighbor sentence representations than the remote sentences in the same paragraph. The intuition behind this constraint comes from a basic idea: \textit{if the $s_j$ is the corresponding sentence for frame $ h_i$, the corresponding sentence for frame $h_{i+1}$ is never before the $s_j$ in sequence}. Specifically, we take the probabilistic sample from the sentence-frame similarity metric. And we propose to apply the differentiable Gumbel-Softmax\cite{gumbel_softmax} tricks to generate predictions and propose three kinds of methods to generate the pseudo-labels that are based on the temporal relation of sentences in the temporal domain: 1) maximum-index sorting; 2) Viterbi algorithm \cite{viterbi}; 3) splitting. Finally, we calculate the Info-NCE contrastive loss based on the pseudo labels in order to guide the network to focus on the fine-grained action matching in sequential videos. 

To evaluate the effectiveness of our weakly supervised video representation method, we conduct extensive experiments on two downstream tasks: video verification in procedures and text-to-video matching. The results of experiments show that our approach outperforms other baselines by a significant margin and also demonstrates the great generalization of our model.  

 We summarize our contributions in three folds:
\begin{itemize}
\item We propose a novel weakly supervised video representation learning pipeline with unaligned text for sequential videos, which can learn powerful and semantic video-text representations.
\item We design multiple granularity contrastive learning loss, including coarse-grained loss and fine-grained loss. Notably, we propose a novel method to implement the temporal alignment between frames and sentences. 
\item Our model also shows strong generalization ability to downstream tasks, such as video sequence verification for procedures in videos and text-to-video matching. 
\end{itemize}

\section{Related Works}\label{sec:related}

\noindent \textbf{Sequential Video}. The same task described in videos may consist of several sequential sub-actions in different orders for a sequential video. Sequential videos are generally accompanied by explanations such as audio or caption. Various kinds of studies related to sequential videos are now in the ascendant. For example, COIN\cite{coin}, Diving\cite{Diving}, CSV\cite{SVIP}, EPIC-KITCHENS\cite{epic-kitchens}, IKEA-ASM\cite{ikea} and Assembly101 \cite{assembly101} provide videos composed by multiple sequential actions and the corresponding step annotations.  \cite{assembly101} proposes a large-scale multi-view video dataset for understanding procedural activities, which is beneficial for the whole community.  \cite{SVIP} defines the pioneering sequence verification task and designs a method based on the alignment of video pairs. However, the method is seriously dependent on video pairs of the same tasks. \cite{procedureactivity} learns to recognize procedural activities in sequential videos with distant supervision\cite{zeng2015distant,mintz2009distant}. \cite{set_supervised} propose an action segmentation method using the set-supervised method for sequential videos.
\cite{kumar2022unsupervised} employs temporal optimal transport to generate pseudo labels to complete joint representation learning and online clustering for sequential video alignment.
D$^3$TW\cite{chang2019d3tw} aligns clips and transcripts with differentiable continuous relaxation.

\noindent \textbf{Vision-text Multi-modality Learning}. Vision-text multi-modality\cite{howto100m, sun2019learning, CLIP, videobert, actbert, LocalVTP, Xie2022Alignment, actionclip, wang2022long} has attracted increasing attention in computer vision communities over the recent year. One of the most representative works is CLIP\cite{CLIP}, which is able to learn a powerful visual representation from natural language supervision with contrastive learning loss. Due to the strong zero-shot generalization ability of the method, a large number of follow-up works have been proposed \cite{X-CLIP, videoclip, LocalVTP, Xie2022Alignment, actionclip, sun2022long,clip-event}. VideoCLIP\cite{videoclip} presents a contrastive approach to pre-train a unified model with video-text pairs. X-CLIP\cite{X-CLIP} effectively expands the pre-trained language-image model to video domains based on a cross-frame attention mechanism. However, these methods heavily rely on strong data augmentation and a large batch size. For downstream tasks, LocVTP\cite{LocalVTP} shows its transfer potentials on localization-based and retrieval-based tasks. CLIP4Clip\cite{clip4clip} uses the pretrained CLIP as our backbone to solve the video clip retrieval task from frame-level input. \cite{bridgeformer} bridges video-text retrieval with multiple-choice questions. LF-VILA\cite{sun2022long} applies a multi-modality temporal contrastive loss to implement long-form video-language pre-training, which heavily relies on the timestamp annotations of clip-sentence pairs.

\noindent \textbf{Video Representation Learning}. Learning good video representations has been heavily investigated in the literature. 3D convolution neural networks (3D-CNNs) are originally considered to learn deep video representations\cite{kinetics, tran2018closer, slowfast}. However, 3D-CNNs are limited to capturing long-term dependencies on the temporal domain with their insufficient receptive field. Due to the ability to capture long-term dependency of the self-attention mechanism\cite{transformer}, vision transformer models\cite{vit, swin, video-swin, MViT, frame-sequence, hu2022transrac, timesformer} show competitive performances against 3D-CNNs in video representation learning. Following the ViT\cite{vit}, many related works emerge. TimeSformer\cite{timesformer} designs different self-attention schemes in the temporal-spatial domain. Video Swin-Transformer \cite{video-swin} adopts the local attention in non-overlapping shifted windows to lead to a better speed-accuracy trade-off. Over recent years, weakly supervised or self-supervised learning\cite{frame-sequence, MPNet} is popular for learning better video representation. Following SimCLR\cite{SimCLR}, \cite{frame-sequence} introduces a self-supervised contrastive transformer framework to learn frame-wise action representations. \cite{cross-modal-rl} proposes a transformer-based cross-modal architecture for zero-shot action recognition. Previous works mainly focus on short-form simple video representation, whereas representation learning of sequential video is underexplored.

\section{Method}\label{sec:method}
\begin{figure*}
\centering
\centerline{\includegraphics[width=\textwidth]{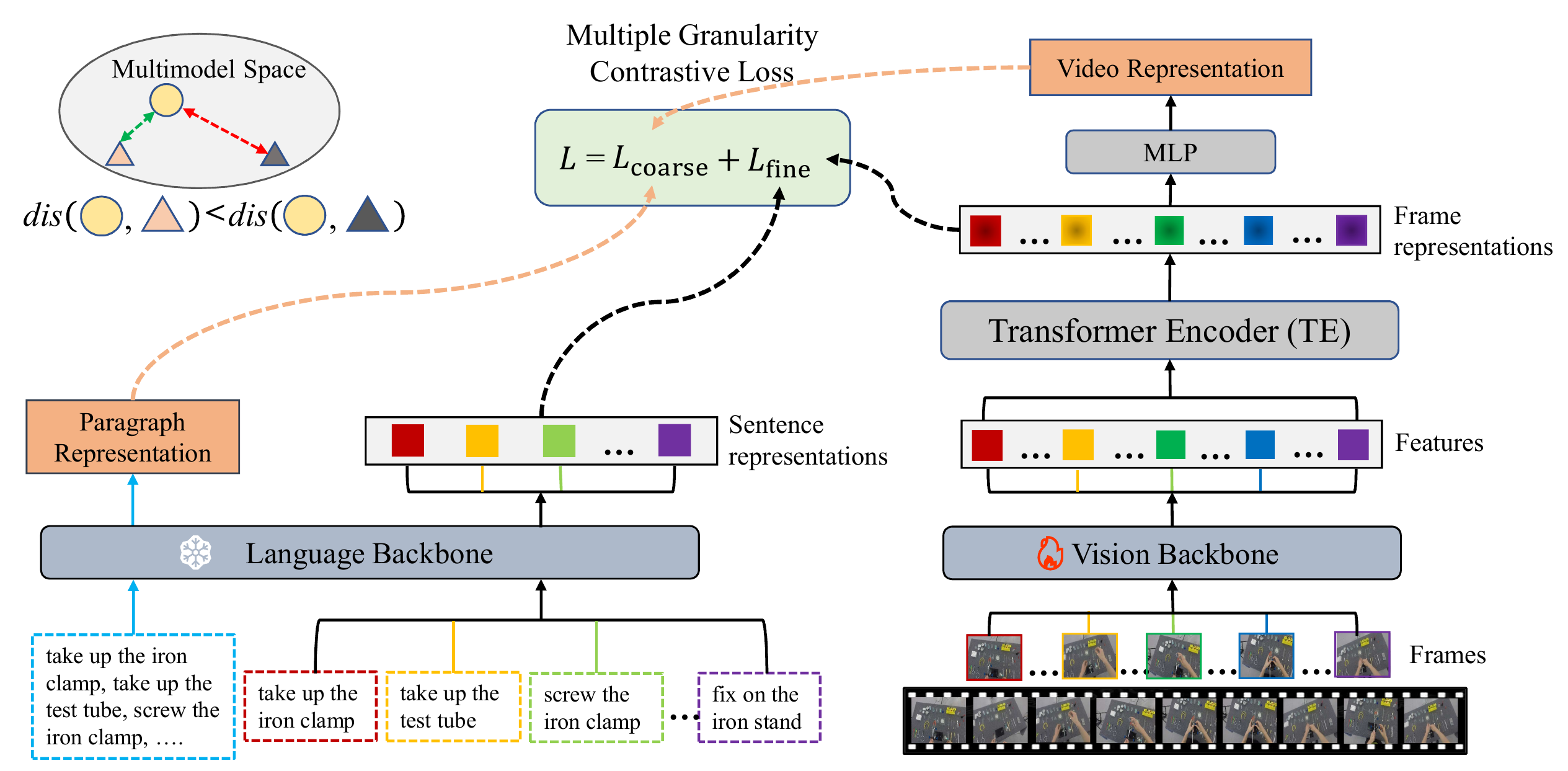}}
\caption{
\textbf{Overview of our framework.}
Our framework consists of three parts: vision representation module, language representation module, multiple granularity loss module. In the vision representation module, we feed the frames sampled from the untrimmed sequential video into the module, then obtain the frame representations and a video representation. In the language representation module, a collection of texts of procedures and the description of the entire video pass into the pre-trained language backbone separately, and we get the sentence representations and a paragraph representation. In addition, we introduce multiple granularity contrastive learning loss to restrict representations in cross-model space.  
}
\vspace{-1.5em}
\label{fig:overview}
\end{figure*}

In this section, we first present the overall architecture of the proposed framework in \cref{sec:overview}. Then we explain the vision representation module and language representation module in \cref{sec:vision-language}, followed by the designed multiple granularity contrastive learning module in \cref{sec:loss1} and \cref{sec:loss2}.

\subsection{Overview}
\label{sec:overview}
\cref{fig:overview} displays the overview of our framework. Our framework consists of three parts: a vision representation module, a language representation module, and a multiple granularity contrastive learning module. In the vision representation module, which shows in the right part of the figure, we sample frames from an untrimmed sequence video as input and extract visual features with the pre-trained vision encoder (unfrozen). After that, we concatenate the visual feature and pass them into the Transformer encoder. The Transformer encoder implements the cross-frame communication with self-attention and outputs the frame representations, following ViT\cite{vit}. Additionally, the results from Transformer encoder are then passed through the MLP module to integrate the frame representations and obtain the video representation. In the left language representation module, a collection of text descriptions of procedures and the description of the entire video pass into the pre-trained language encoder (frozen) separately, then we can obtain the sentence representations and a paragraph representation. More explanation about the aforementioned modules is in \cref{sec:vision-language}. Finally, we introduce multiple granularity contrastive loss to restrict learned representations in cross-model space.

\subsection{Vision-Language Modules}
\label{sec:vision-language}
As illustrated in \cref{fig:overview}, multi-level video representation and language representation are produced by the vision module and language module, respectively.

\noindent \textbf{Vision module}.
Following\cite{TSN}, given an untrimmed sequence video, we uniformly split the raw video into $N$ clips and randomly sample one frame per clip to form a sequence of $N$ frames, $X=\{x_1, x_2, \dots, x_N\}$. Then we feed the frame sequence $X$ into the pre-trained vision encoder $E_v$ to produce a sequence of feature maps $\{f_1, f_2, \dots, f_N\}$. This process can be denoted as $f_i = E_v(x_i), i\in [1, 2, \dots, N]$. After that, we prepend a learnable embedding $x_{cls}$ to the sequence of features, called $[class]$ token \cite{vit}. Then as \cref{eq:framerepresentations} shown, our method learns the frame representations by utilizing the transformer encoder (TE) to embed temporal and context information into frame representations $H =\{h_1, h_2, \dots, h_N\}$. 
 
\begin{equation}
    H = \text{TE}(\left[x_\text{cls}, f_1, f_2, \dots, f_N\right] + e^\text{pos})  \\
\label{eq:framerepresentations}
\end{equation}
where  $[. , .]$ concatenates the features of frames and $[class]$ token. And $e^\text{pos}$ represents the temporal position embedding of sequence.

At last, the MLP module, which consists of a full connection layer, takes all frame representations $H$ as input and outputs a video representation $v$ as follows:
\begin{equation}
    v  = \text{MLP}(H)  \\
\label{eq:videorepresentation}
\end{equation}

\noindent \textbf{Language module}. Specifically in our model, given a sequence of $K$ text descriptions of procedures $ T = \{t_1, t_2, \dots, t_K\}$, we first feed individual procedure texts into the frozen pre-trained language encoder $E_l$ to produce sentence representations $S=\{s_1, s_2, \dots, s_K\}$. The process can be denoted as $s_i = E_l(t_i), i \in [1,2,..K]$.

In the meantime, we combine the sequence of text descriptions of procedures $ T $ into a single text description of the entire video. Then, the pre-trained language encoder $E_l$ extract a paragraph-level language representation $l$ as follows:
\begin{equation}
    l =E_l(\left[t_1, t_2, \dots, t_K\right]) \\
\label{eq:paragraphrepresentation}
\end{equation}
where  $[. , .]$ represents simply the sequential concatenation of strings.
\vspace{-5pt}
\subsection{Coarse-grained Contrastive Loss}
\label{sec:loss1}

We first conduct contrastive learning at the video-paragraph level. Specifically, through the vision-language module that is explained in \cref{sec:vision-language}, we obtain a video representation $V$ and paragraph representation $L $, where $V, L \in \mathbb{R}^{1 \times D}$. Then use one batch of data, $V = \{v_1,v_2,\dots,v_N\}$, $L =\{l_1,l_2,\dots,l_N\}$, to calculate the loss.

After that, we formulate the global video-paragraph alignment into the standard contrastive framework\cite{CLIP} based on InfoNCE loss\cite{InfoNCE} as follows:
\begin{equation}
    L_{\text{InfoNCE}}(V,L)=-\frac{1}{N} \sum_{i=1}^{N} \log\frac{\exp{(\varphi(v_i,l_i)/\tau)}}{\sum_{j=1}^{N} \exp{(\varphi(v_j,l_j)/\tau})} \\
\label{eq:infonce}
\end{equation}
\vspace{-10pt}
\begin{equation}
    \varphi(v_i,l_i) = \frac{v_i}{\left\lVert v_i \right\rVert} \cdot \frac{l_{i}^{T}}{\left\lVert l^T \right\rVert}
\label{eq:cos_sim}
\end{equation}
where $\tau$ is the temperature parameter optimized during training\cite{CLIP}. And $\varphi(.,.)$ represents the cosine similarity function, and $N$ is the number of video-text pairs. The $L_\text{InfoNCE}$ represents the InfoNCE loss.

Last, as shown \cref{eq:global}, we calculate symmetrically video-text and text-video loss by \cref{eq:infonce} to obtain the coarse-grained contrastive loss $L_\text{coarse}$:
\begin{equation}
    L_\text{coarse}= L_{\text{InfoNCE}}(V,L) + L_{\text{InfoNCE}}(L,V) \\
\label{eq:global}
\end{equation}

Showing in the upper left of \cref{fig:overview}, the coarse-grained global contrastive loss $L_\text{coarse}$ restricts the representation in the cross-model latent space with video-paragraph level supervision.

\subsection{Fine-grained Contrastive Loss} \label{sec:loss2}



\begin{figure}[ht!]
  \centering
    \begin{subfigure}{0.23\textwidth}
      \centering   
      \includegraphics[width=\textwidth]{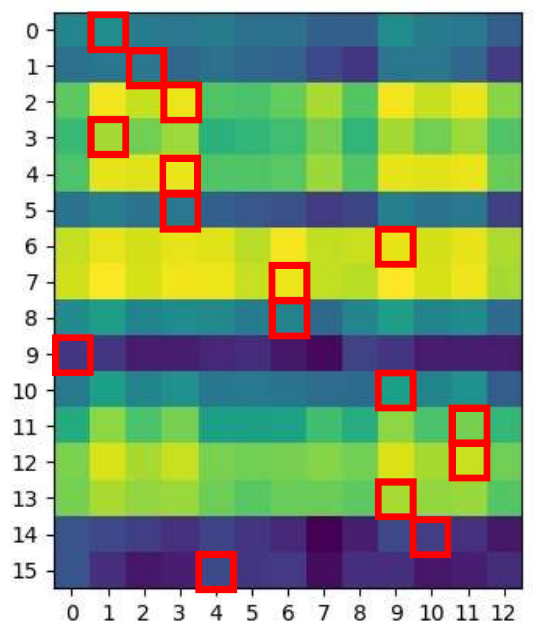}
        \caption{The Output of Gumbel-Softmax}
        \label{fig:1a}
    \end{subfigure}        
    \hfill 
    \begin{subfigure}{0.23\textwidth}
      \centering   
      \includegraphics[width=\textwidth]{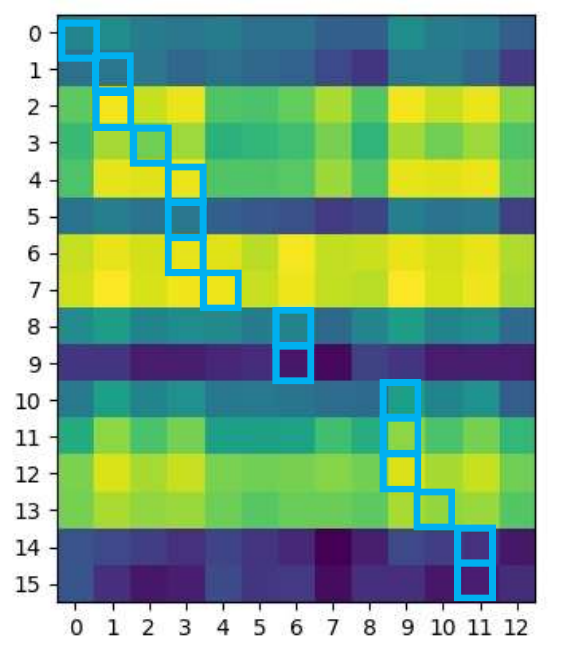}
        \caption{Maximum-index Sorting}
        \label{fig:1b}
    \end{subfigure}
    \hfill
    \begin{subfigure}{0.23\textwidth}
      \centering   
      \includegraphics[width=\linewidth]{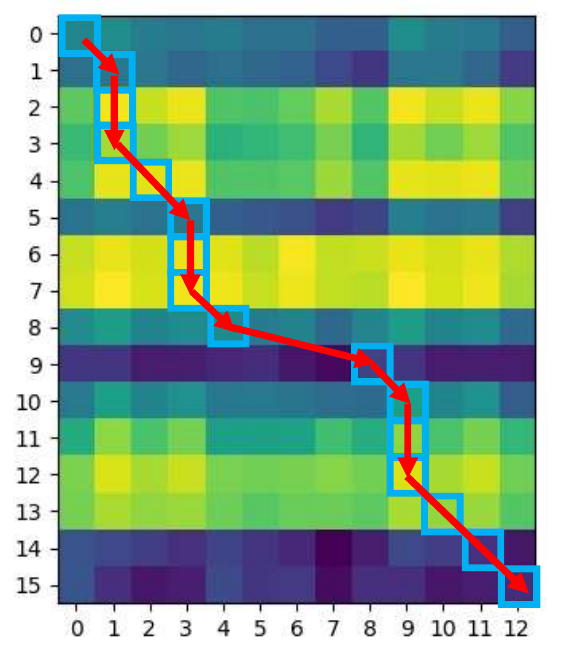}
        \caption{Viterbi Algorithm}
        \label{fig:1c}
    \end{subfigure}
    \hfill
    \begin{subfigure}{0.23\textwidth}
      \centering   
      \includegraphics[width=\linewidth]{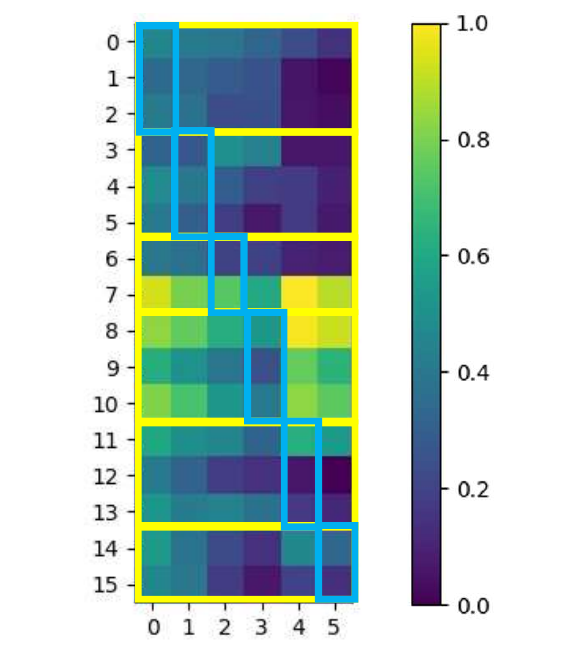}
        \caption{Splitting}
        \label{fig:1d}
    \end{subfigure}
\caption{
\textbf{Visualization} of fine-grained contrastive loss. The upper left figure shows the similarity matrix with Gumbel-Softmax. The other three figures show three kinds of pseudo-labels generation methods respectively: 1) maximum-index sorting; 2) Viterbi algorithm; 3) splitting.
}
\label{fig:gumbel}
\vspace{-1em}
\end{figure}
Due to the lack of frame-level annotations, there is no annotation to locate the start frame and end frames per action. A frame can't know its correct corresponding sentence.
To overcome this problem, we propose an essential hypothesis based on the temporal relation between sentences and frames: if  $s_j$ is the corresponding sentence representation for the frame representation $ h_i$, the sentence representation for frame representation $h_{i+1}$ should be in the set of  $\{s_j,s_{j+1},s_{j+2},\dots, s_{K} \}$ and never in the set of  $\{s_1,s_{2},\dots, s_{j-1} \}$. The visualization of fine-grained contrastive loss can be seen in \cref{fig:gumbel}.

Specifically, in our model, we first obtain the sequence of frame-level sequences of representations $H$ and sequence of sentence-level representations $S$ through the vision-language module. As \cref{eq:fine2} shows, we symmetrically calculate the fine-grained contrastive learning loss, named $L_{fine}$, to achieve frame-sentence alignment. 
\begin{equation}
    \begin{aligned}
        L_{fine} = &\textit{CE}(\psi_{\text{preds}}(H,S), \phi_{\text{pseudo}}(H,S)) \\
                    &  +  \textit{CE}(\psi_{\text{preds}}(S,H), \phi_{\text{pseudo}}(S,H)) \\
    \end{aligned}
    \label{eq:fine2}
\end{equation}
where \textit{CE} is the Cross-Entropy loss. We use $\psi_{\text{preds}}$ to predict the most related sentence $s_j$ with one frame, where $s_j \in S$. The $\phi_{\text{pseudo}}$ could utilize the probability distribution of prediction and the similarity matrix of $H$ and $S$ to generate the pseudo labels as ground truth.
Then as \cref{eq:fine2} shown,  we calculate $L_\text{fine}$ by the \textit{CE} loss of the prediction and pseudo labels. 
And we separately introduce two methods of $\psi_{\text{preds}}$ and $\phi_{\text{pseudo}}$ in \cref{sec:sorting} and \cref{sec:viterbi}. The method of Gumbel-Softmax with splitting is shown in \cref{sec:ablation_pseudo}.

\subsubsection{Gumbel-Softmax with Sorting} \label{sec:sorting}

We first use \cref{eq:cos_sim} to calculate the similarity matrix between the frame representations $H$ and its sentence representations $S$. And we obtain the first prediction by \cref{eq:preds_gumbel}.
\vspace{-0.5em}
\begin{equation}
    \psi_{\text{preds}}(H,S) = \text{Gumbel-Softmax}(\varphi(H,S))
    \label{eq:preds_gumbel}
\end{equation}
where Gumbel-Softmax is the straight-through Gumbel-Softmax function\cite{gumbel_softmax}. We utilize the Gumbel-Softmax to ensure the dispersed sampling from the original distribution can be calculated for the gradients in the backward pass.
Then, we get the maximum index through $\arg\max$ and sort the maximum-index list to an increasing order to generate pseudo labels. We regard them as the ground truth, which shows in \cref{fig:1b} in blue.
Finally, we finish the first kind of $\psi_{\text{preds}}$ and $\phi_{\text{pseudo}}$  by \cref{eq:preds_gumbel,eq:pseudo_gumbel}.
\begin{equation}
\phi_{\text{pseudo}}(H,S) = \text{sort} \left[ \mathop{\arg\max}_{i\in[1,K]}(\psi_{\text{preds}}(H,S)_{N\times K}) \right] 
\label{eq:pseudo_gumbel}
\end{equation}

\subsubsection{Gumbel-Softmax with Viterbi} \label{sec:viterbi}

Following the Viterbi algorithm\cite{viterbi}, it could generate the maximum a posteriori probability estimate, called the Viterbi path. The original Viterbi algorithm needs two important matrices: transition matrix and emission matrix. As shown in \cref{eq:viterbi_emission}, we use the similarity of the language and vision features as the emission with the shape $[N, K]$, where $N$ means the number of sampled frames and $K$ is the total number of its labels. Specifically in our method, as \cref{eq:viterbi_transition}shows, we use one upper triangular mask matrix as the transition matrix to limit the path of probability, which could make sure the way won't go back. Based on the Viterbi path (shown in \cref{fig:1c}), we obtain the pseudo-labels by \cref{eq:viterbi}. Different from our method using Viterbi algorithm to generate pseudo labels, \cite{review1_1,review1_3} apply Viterbi decoding prediction, and activities have constant action orders. More details about the Viterbi algorithm can be seen in supplementary materials. 
\vspace{-1em}
\begin{equation}
    \text{Transition matrix:} A=
    \begin{bmatrix}
        \frac{1}{n} & \dots  & \frac{1}{n} \\
               & \ddots & \vdots \\
        0      &        & \frac{1}{n}\\
    \end{bmatrix}_{N \times N}
    \label{eq:viterbi_transition}
\end{equation}
\vspace{-10pt}
\begin{equation}
    \text{Emission matrix: } B=\varphi(H,S)
    \label{eq:viterbi_emission}
\end{equation}
\vspace{-10pt}
\begin{equation}
    \phi_{\text{pseudo}}(H,S) = \text{Viterbi}(A, B) 
    \label{eq:viterbi}
\end{equation}

\subsubsection{Training Loss}

In conclusion, we train our module with the combination of the  proposed coarse-grained contrastive loss and fine-grained contrastive loss:
\vspace{-0.5em}
\begin{equation}
    L= L_{\text{coarse}} + \lambda_1 L_{\text{fine}} \\
\label{eq:lossVR}
\vspace{-0.5em}
\end{equation}
where $\lambda_1$ represents the weight of fine-grained contrastive loss.

\section{Experiments}
\label{sec:experiments}

In this section, we first introduce the implementation details, evaluation benchmarks and evaluation metrics in \cref{sec:datasets}. The experiments to verify the effectiveness of baselines for video-text representation learning are shown in \cref{sec:evaluation}. In addition, we also transfer our proposed framework to downstream sequence verification in \cref{sec:Verification} and text-to-video matching tasks in \cref{sec:matching}. 

\subsection{Experimental Details}
\noindent \textbf{Implementation Details}. 
\label{sec:implementation}
The vision backbone we employ is the pre-trained CLIP vision encoder based on ViT-B\cite{vit}. And the model is initialized adopting Kaiming and Xavier uniform initialization for different layers\cite{kaiming,MAE}. In our module, the parameter of the vision backbone is unfrozen and finetuned when training. On the other hand, the language backbone is the pre-trained CLIP text encoder whose parameter is frozen totally. More implementation details can be seen in supplementary materials. 

\noindent \textbf{Datasets}. \label{sec:datasets}
We conduct experiments on the datasets COIN-SV, Diving-SV and CSV. COIN-SV is rearranged from COIN and composed of 36 tasks that contain more than 20 comprehensive instructional videos in the training dataset. Diving-SV is rearranged from Diving and contains 48 kinds of diving competition videos. And CSV\cite{SVIP} includes 45 procedures for training and 25 procedures for testing. 
In these datasets, all kinds of videos in the test set are unseen in the training set.

\noindent \textbf{Testing phase}.
During inference, we apply the method that distinguishes positive pairs from negative pairs to evaluate the quality of learned video representations. Specifically in this paper, we calculate the normalized Euclidean distance between two video representations $v_1$ and $v_2$ in the same video pair:
\begin{equation}
    d = dis(v_1,v_2)  \\
\end{equation}
\vspace{-10pt}
\begin{equation}
    y=\begin{cases}
    1, d \leq \tau \\
    0, otherwise \\
    \end{cases}
\label{eq:logist}
\end{equation}
where $dis(.,.)$ means the $\ell2$-normalization Euclidean distance function. $\tau$ is a threshold to decide whether the sequences are consistent. $y = 1$ means the two sequences of videos are consistent, otherwise inconsistent.

\noindent \textbf{Evaluation Metrics}. 
We adopt the Area Under ROC Curve (\textbf{AUC}) as the measurement for all of our experiments, which is commonly used to evaluate the performance in the field of anomaly detection \cite{mist} and face verification\cite{arcface}. Higher AUC means better performance.

\subsection{Comparison of baselines}
\label{sec:evaluation}
Under weak supervision, the only annotations we know are the text descriptions of procedures, but the timestamps of actions and video task classification are unknown. The results of weakly supervised video sequence verification are shown in \cref{tab:baseline}. We compare our method with other baselines, including 1) MIL-NCE\cite{MIL-NCE}. 2) CAT, we change the SVIP\cite{SVIP} model architecture and add a text encoder to adapt to this task. 3) VideoSwin+MLP, we adopt the video swin transformer\cite{video-swin} as the vision encoder to extract frame features. 4) CLIP+Transformer Encoder+Pool. 5) CLIP+Transformer Encoder+MLP. To adapt to the task, we apply the CLIP text encoder as the text encoder of baselines except for MIL-NCE. Other methods but ours only calculate the coarse-grained contrastive loss.

\begin{table}[ht!]
    \centering
    \resizebox{\linewidth}{!}{
    \begin{tabular}{c|c|c|c|c}
    \hline
    \multirow{2}[4]{*}{Method} & \multirow{2}[4]{*}{Text Encoder} & \multicolumn{3}{c}{Weakly Supervised (w/o CLS)} \bigstrut\\
\cline{3-5}          &       & CSV   & Diving-SV & COIN-SV \bigstrut \\
    \hline
    MIL-NCE\cite{MIL-NCE} & MLP\cite{MIL-NCE}   & 53.02  & 58.49  & 47.95 \\
    \hline
    CAT\cite{SVIP} & \multirow{4}[1]{*}{CLIP\cite{CLIP}} & 70.63  & 77.87  & 47.70  \\
    VideoSwin\cite{video-swin}+MLP &       & 62.48  & 60.88  & \textbf{54.73}  \\
    CLIP\cite{CLIP}+TE\cite{vit}+Pool &       & 58.67  & 72.13  & 49.79  \\
    CLIP\cite{CLIP}+TE\cite{vit}+MLP &       & 74.82  & 81.47 & 50.13  \\
    \hline
    \textbf{Ours}  &   CLIP\cite{CLIP}    & \textbf{79.80}  & \textbf{85.19} & 52.56 \\
    \hline
    \end{tabular}%
    }\vspace{-0.5em}
	\caption{Results of representation learning for weakly supervised video sequence verification task. 
	}

\label{tab:baseline}
\vspace{-1.0em}
\end{table}

The results in \cref{tab:baseline} demonstrate that multiple granularity contrastive learning is effective for learning discriminative video representations under weak supervision.

\subsection{Sequence Verification}
\label{sec:Verification}

\begin{table}[htbp]
\centering
\resizebox{\linewidth}{!}{
    \begin{tabular}{c|c|c|c|c}
    \hline
    \multirow{2}[4]{*}{Method} & \multirow{2}[4]{*}{Pre-train} & \multicolumn{3}{c}{Supervised (w CLS)} \bigstrut\\
\cline{3-5}          &       & CSV   & Diving-SV & COIN-SV \bigstrut\\
    \hline
    MIL-NCE\cite{MIL-NCE} & HowTo100M\cite{howto100m} & 56.16  & 63.43  & 47.80  \\
    Swin\cite{swin}  & K-400\cite{kinetics} & 54.06  & 73.10  & 43.70  \\
    TRN\cite{TRN}   & K-400\cite{kinetics} & 80.32  & 80.69  & 57.19  \\
    CAT\cite{SVIP}   & K-400\cite{kinetics} & 83.02  & 83.11  & 51.13  \\
    CLIP\cite{CLIP}+TE\cite{vit}+MLP & CLIP\cite{CLIP}  & 79.38  & 83.48 & 48.50  \\
    \hline
    Ours (weakly supervised) & CLIP\cite{CLIP}  & 79.80  & 85.19 & 52.56  \\
    \hline
    \textbf{Ours}  & CLIP\cite{CLIP}  & \textbf{86.92}  & \textbf{86.09} & \textbf{59.57} \\
    \hline
    \end{tabular}%
    }
    \vspace{-0.5em}
\caption{Results of downstream video sequence verification task under supervised setting.}
\label{tab:sequence_verification}
\vspace{-1.0em}
\end{table}


Following the setting of sequence verification \cite{SVIP}, we know the classification of videos but yet do not know the timestamp of actions. 
The testing results on sequence verification compared to other methods are shown in the \cref{tab:sequence_verification}. 
We can use class information for sequence verification of procedures in videos.

For a fair comparison, some adjustments have been made to the architecture of our model in this task setting. 
Specifically, we add a classification layer on the top of the video representation and the classification loss to our model. Besides, we apply the adjusted video sequence alignment mechanism by ours and train with pair data that are the same as SVIP\cite{SVIP}.
This adjusted model is named "Ours". In addition, we also compare with some state-of-the-art methods\cite{SVIP,swin,TRN} of sequence verification and change some video-language pre-trained model \cite{howto100m} accordingly to adapt to this task. Weakly supervised means no classification information of tasks.
To clarify the improvements from technical differences, we replace the visual backbone of CAT\cite{SVIP} with CLIP-ViT\cite{CLIP} to form CLIP+TE+MLP. Then, we improve performance by adjusting network structure, e.g., the position of SEQ loss.   
Observed \cref{tab:sequence_verification}, our model outperforms them by a notable margin on all the considered datasets. The results also demonstrate that the fine-grained contrastive loss we proposed enforces the model to learn more discriminative representations.  The results of our weakly supervised model, which  surpasses other supervised methods, demonstrate our model's excellent performance.

\subsection{Text-to-Video Matching}
\label{sec:matching}
\noindent \textbf{Setting}.
We validate the performance of the video-language representations on text-to-video matching, which aims to find the correct video corresponding to a sequence of texts from a series of videos. Specifically, we train our model on the CSV dataset under weak supervision and test it on our proposed benchmark about text-to-video matching. 
This task evaluates the model's ability to learn semantic and generalized video representations.

\noindent \textbf{Benchmark}.
To better evaluate the text-to-video matching, we rearrange the test set of CSV\cite{SVIP} and propose a new scripted benchmark, named \textbf{CSV-Matching}. It has 800 text-video pairs. Each text-video pair is composed of one sequence of text descriptions of procedures and five videos. All of the videos describe the same task but hold different procedures. There is only one correct video matching the text descriptions in each pair. More details about the text-to-matching benchmark will show in the supplementary materials.

The text-to-video matching results in \cref{tab:matching} indicate  that our method has the best performance. And due to data of CSV-Matching being unseen when training, it shows that our method has a more powerful generalization ability. 

\begin{table}[htbp]
\centering
    \begin{tabular}{c|c}
    \hline
    \multirow{2}[3]{*}{Method } & Text-to-Video  Matching \bigstrut \\
\cline{2-2}          & CSV-Matching \bigstrut\\
    \hline
    MIL-NCE\cite{MIL-NCE} & 60.02         \\
    CAT\cite{SVIP}   & 53.54 \\
    CLIP\cite{CLIP} +TE\cite{vit} +MLP  & 62.67 \\
    \hline
    \textbf{Ours}  &  \textbf{65.23} \\
    \hline
    \end{tabular}%
\vspace{-0.5em}
\caption{Results of text-to-video matching task on our proposed benchmark \emph{CSV-Matching}. We evaluate the results using AUC. }
\label{tab:matching}
\vspace{-1.0em}
\end{table}

\section{Analysis}\label{sec:ablation}
In this section, we first analyze the impact of different backbones in \cref{sec:ablation_backbone}.
conduct comprehensive ablation studies of multiple granularity contrastive loss and pseudo-label generation in \cref{sec:ablation_loss,sec:ablation_pseudo}.
Moreover, we analyze our limitations and broader impact. 

\subsection{Ablation of Backbone}

\label{sec:ablation_backbone}
As \cref{tab:backbone} shown, our method based on CLIP-ViT obtains the best performance compared with other backbones. In addition, results indicate that fine-grained and multi-grained losses improve performance under weak supervision and supervision, respectively.
\begin{table}[htbp]
\centering

\resizebox{\linewidth}{!}{
\begin{tabular}{c|c|c|c|c|c|c}
\hline
\multirow{2}[4]{*}{Backbone} & \multirow{2}[4]{*}{Pretrained} & \multicolumn{3}{c|}{Weakly Supervised (w/o CLS)} & \multicolumn{2}{c}{Supervised (w CLS)} \bigstrut\\
\cline{3-7}           &            & $L_{\text{coarse}}$     & $L_{\text{fine}}$       & CSV        & $L_{\text{coarse}}$ +$L_{\text{fine}}$ & CSV \bigstrut\\
\hline
\multirow{2}[2]{*}{ResNet50\cite{ResNet}} & \multirow{2}[2]{*}{ImageNet-1K} & \multirow{2}[2]{*}{\ding{51}} & \graycross & 76.22      & \graycross & 78.83  \bigstrut[t]\\
           &            &            & \ding{51}  & 78.32      & \ding{51}  & 81.00  \bigstrut[b]\\
\hline
\multirow{2}[2]{*}{ViT-B/32\cite{vit}} & \multirow{2}[2]{*}{ImageNet-21K} & \multirow{2}[2]{*}{\ding{51}} & \graycross & 73.88      & \graycross & 81.66  \bigstrut[t]\\
           &            &            & \ding{51}  & 75.18      & \ding{51}  & 82.11  \bigstrut[b]\\
\hline
\multirow{2}[2]{*}{CLIP-ViT\cite{CLIP} (Ours)} & \multirow{2}[2]{*}{Text-Image Pair} & \multirow{2}[2]{*}{\ding{51}} & \graycross & 78.49      & \graycross & 83.58  \bigstrut[t]\\
           &            &            & \ding{51}  & 79.80      & \ding{51}  & 86.92  \bigstrut[b]\\
\hline
\end{tabular}%

}
\vspace{-0.5em}
\caption{Results of our method with different backbone on CSV.}
  \label{tab:backbone}%
  \vspace{-1.0em}
\end{table}%
\subsection{Ablation of Multiple Granularity Contrastive Loss}
\label{sec:ablation_loss}

In this section, we conduct comprehensive ablation studies to investigate the effects of our multiple granularity contrastive loss.
To better demonstrate the superiority of our method, we present the loss ablation experiments on the sequence verification task under supervision with classification in \cref{tab:loss}. As shown, both coarse-grained contrastive loss $L_{\text{coarse}}$ and fine-grained loss $L_{\text{fine}}$ are crucial. Specifically, the method with coarse-grained and fine-grained contrastive loss surpasses the method without them by 3.34\%. 
    

\begin{table}[htbp]
  \centering
  \
    \begin{tabular}{c|cc|c}
    \hline
    Method & $L_{\text{fine}}$ & $L_{\text{coarse}}$ & CSV \\
    \hline
    \multirow{4}[2]{*}{Ours (w CLS)} & \graycross & \graycross & 83.58 \\
          & \ding{51} & \graycross & 84.85 \\
          & \graycross & \ding{51} & 84.32 \\
          & \ding{51} & \ding{51} & \textbf{86.92} \\
    \hline
    \end{tabular}%
    \vspace{-0.5em}
    \caption{Ablation studies of our proposed multiple granularity contrastive loss on CSV. To verify the effectiveness of $L_{\text{fine}}$ and $L_{\text{coarse}}$ separately, we conduct experiments on video verification task.}
  \label{tab:loss}%
  \vspace{-1.0em}
\end{table}%

Introducing the fine-grained loss $L_{\text{fine}}$ brings 2.6\% performance improvement compared to only using coarse-grained contrastive loss $L_{\text{coarse}}$. Comparing only uses $L_\text{coarse}$ or uses  $L_\text{fine}$, the result indicates that the model training with more fine-grained information is better than coarse information. 
By restricting the video representation to frame-sentence level latent space, the fine-grained contrastive loss can help the model learn more discriminative video representations.

\begin{figure}[ht!]
 \centering    
      \includegraphics[width=0.45\textwidth]{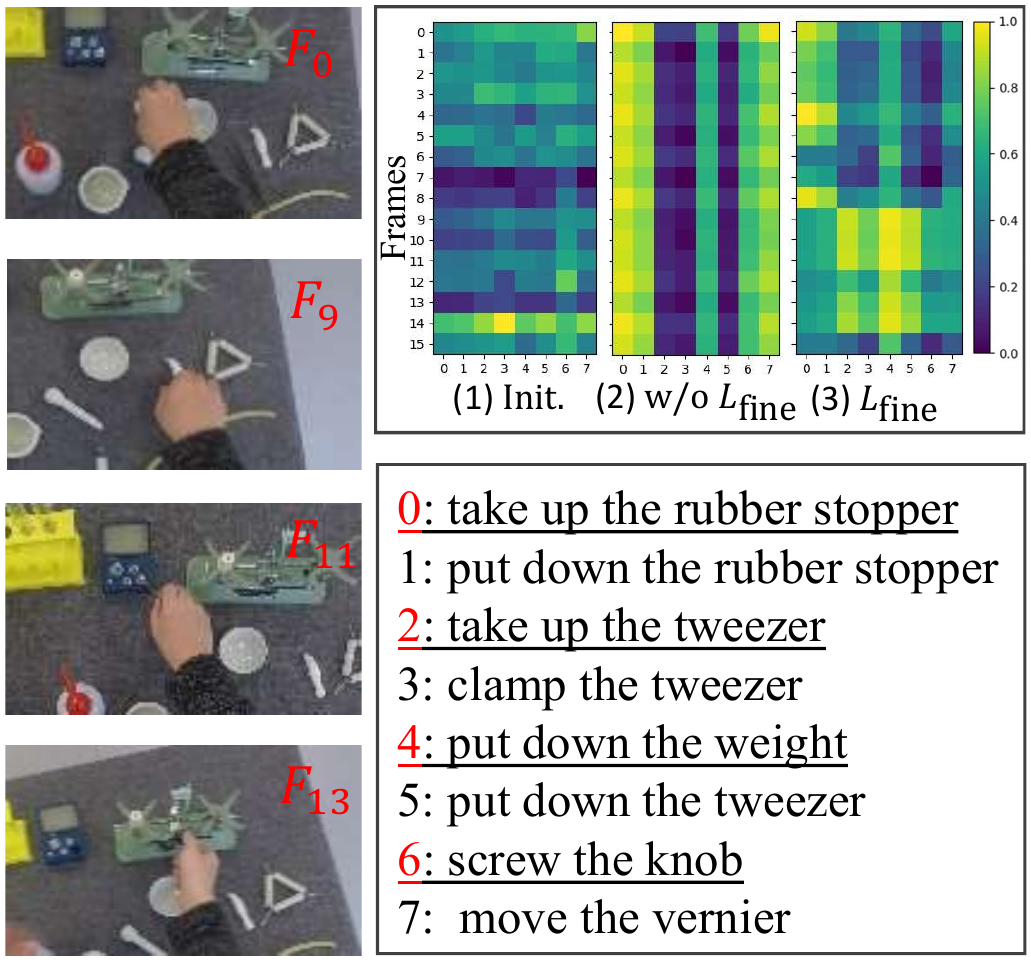}
\caption{\textbf{Visualization} of ablation study about fine-grained contrastive loss.}
    \label{fig:vis}
\end{figure}





\vspace{-1.0em}
The visualization of the ablation study about fine-grained loss, as shown in \cref{fig:vis}, illustrates fine-grained contrastive loss implements the alignment between frames and sentences. 

\subsection{Ablation of the Pseudo-Label Generation}
\label{sec:ablation_pseudo}

\noindent \textbf{Splitting.} 
Splitting means we split the sequence of frame representations or sentence representations uniformly into several parts to keep the sequence length of frame representations or sentence representations equal.
The values belonging to the same part will be added and then averaged. After that, we get a square matrix and output probability distribution of prediction. The elements along the diagonal are regarded as pseudo labels. Then calculate the fine-grained contrastive loss as \cref{eq:fine2}. This process is shown in \cref{fig:1d}, and the blue boxes represent the pseudo-labels.
\begin{table}[htbp]
  \centering
    \begin{tabular}{c|c|c|c}
    \hline
    Method & $L_{\text{fine}}$ & Pseudo-label generation & CSV \\
    \hline
    \multirow{4}[4]{*}{Ours} & \graycross & \graycross & 74.82 \\
    \cline{2-4}          & \multirow{3}[2]{*}{\ding{51}} & split & 72.75  \\
          &       & viterbi & 78.46  \\
          &       & \textbf{sort}  & \textbf{79.80} \\
    \hline
    \end{tabular}%
    \vspace{-0.5em}
  \caption{Ablation studies of the type of pseudo-label generation on our proposed method.}
  \label{tab:gumbel}%
  \vspace{-1.0em}
\end{table}%



\vspace{-0.2em}
We conduct ablation studies about three methods of pseudo-label generation in the fine-grained loss $L_\text{fine}$ showing in \cref{tab:gumbel}. Specifically, we validate the effectiveness of different kinds of coarse-grained contrastive loss on the weakly supervised video verification task. 
The results show the algorithms of maximum-index sorting and Viterbi are performing better than splitting. The method of splitting matrices into several parts and aligning sequences along the diagonal is too simple and crude .

\noindent \textbf{Broader Impact and Limitations.}
\label{sec:limitation}
In realistic sequential videos, sub-actions could be repeated. It could mislead the model to generate biased pseudo-labels and lead to the deterioration of performance.More analysis can be seen in  supplementary materials. Moreover, the proposed method will likely be applied to behavior detection, healthcare, online education, industrial generation, etc. 


\section{Conclusions}\label{sec:conclusions}
In this paper, we propose a novel framework of weakly supervised video representation learning for sequential videos. Borrowing the multi-modal contrastive learning from CLIP, our method can learn video representation with unaligned text and video without relying on the accurate time-stamp level text-video annotation.
We propose a multiple granularity loss where the video-paragraph contrastive loss constrains the matching between the whole video and the complete script, and a fine-grained frame-sentence contrastive loss constrains the matching between each action and its descriptions. We also propose to generate pseudo labels with temporal consistency in video and text. 
Experiments results show that our design is effective, and our method achieves state-of-the-art performance when transferred to downstream video sequence verification and text-to-video matching tasks.

\noindent \textbf{Acknowledgements}. 


The work was supported by National Key R\&D Program of China (2018AAA0100704), NSFC \#61932020, \#62172279,  Science and Technology Commission of Shanghai Municipality (Grant No. 20ZR1436000), and “Shuguang Program" supported by Shanghai Education Development Foundation and Shanghai Municipal Education Commission.

\newpage

\appendix


\twocolumn[
\centering
\Large
\textbf{Weakly Supervised Video Representation Learning \\with Unaligned Text for Sequential Videos} \\
\vspace{0.5em}Supplementary Material \\
\vspace{1.0em}
] 
\appendix

\section{Extra Experiment Studies}
In this section, we present additional ablation studies about our method, including the effects of batch size, the number of clips sampled per video, the approach of extracting paragraph-level language representation, and sequence align loss.

\subsection{Implementation Details}
We implement our method with PyTorch. The vision backbone we employ is the pre-trained CLIP vision encoder based on ViT-B\cite{vit}. And the model is initialized adopting Kaiming and Xavier uniform initialization for different layers\cite{kaiming,MAE}. In our module, the parameter of the vision backbone is unfrozen and finetuned when training. On the other hand, the language backbone is the pre-trained CLIP text encoder whose parameter is frozen totally. We split the raw video into 16 clips for a sequential video and randomly sample one raw frame from each clip in the training and uniformly sample frames in inference. The projection layer adopts a fully connected layer. The hidden layer dimension of transformer encoder\cite{vit} is 1024, and the depth is 2. The dimension of the video representations and paragraph representations is 512. The $\lambda_1$ in our model is equal to 1. 
The experiments are conducted on 4 NVIDIA 2080Ti GPUs with batch size 8. We adopt an AdamW optimizer\cite{adamw} with cosine annealing learning rate scheduler with a base learning rate of  $5\times 10^{-4}$, and weight decay 0.01. More implementation details should be seen in the supplementary materials.
And expect the experiment of sequence align loss to be conducted on the supervised sequence verification task, other experiments are conducted on the weakly supervised sequence verification task. We conduct all experiments on CSV dataset.

\subsection{Batch size}
To adapt to the change in batch size, we increase or decrease the learning rate exponentially. As \cref{tab:batchsize} shown, our method achieves the best performance when the batch size is equal to 16.
The larger the batch size, the more likely multiple videos of the same task will appear in the same mini-batch.
Due to the limitation of GPU memory, the largest batch size can be set as 8 if we unfreeze the vision backbone.
\begin{table}[htbp]
  \centering

    \begin{tabular}{c|c|c|c}
    \hline
    Method & Batch size & Frames & CSV \\
    \hline
    \multirow{5}[2]{*}{Ours} & 4     & 16    & 65.94 \\
          & 8     & 16    & 67.46 \\
          & 16    & 16    & 69.42 \\
          & 24    & 16    & 69.21 \\
          & 32    & 16    & 69.16 \\
    \hline
    \end{tabular}%
    \caption{Ablation studies of batch size on our proposed method}
    \label{tab:batchsize}%
\end{table}%
\subsection{Sampling}

While changing the frames of sampling per video, the training time is doubled with the increasing number of frames. In this ablation study, all models have been training no more than 100 epochs or 12 hours on two GPUs due to the limitation of computing resources. 

As \cref{tab:num_clip} shown, when frames are set to 16, our model achieves the best performance. It is worth noting that, with more training steps (about twice the training time), the performance of  $32$ frames will increase to 68.20. However, we choose $16$ as the default frame to balance batch size, number of frames, and training cost, we choose $16$ as the default frame. 

The significant reason for choosing sampling frames rather than video clips is the limitation of computational resources. Fine-tuning the full pre-trained backbone, such as VideoCLIP\cite{videoclip}, is expensive. Similarly, to balance the efficiency of the network and fairly compare our method with CAT\cite{SVIP}, we choose 16 frames as the same as CAT.

\begin{table}[htbp]
  \centering

    \begin{tabular}{c|c|c|c}
    \hline
    Method & Batch size & Frames & CSV \\
    \hline
    \multirow{4}[2]{*}{Ours} & 8     & 8     & 58.43  \\
          & 8     & 16    & 67.46  \\
          & 8     & 32    & 65.64  \\
          & 8     & 48    & 64.40  \\
    \hline
    \end{tabular}%
      \caption{Ablation studies of the number of frames.}
  \label{tab:num_clip}%
\end{table}%

\subsection{Paragraph feature}
We design two ways to extract the feature of the paragraph. The one is concatenating all sentences into a paragraph description. Then we can obtain the paragraph-level representation by feeding the paragraph description into the language encoder. The other method is that feed individual procedure texts into the frozen language encoder to produce sentence representations and then obtain a paragraph-level representation by temporal mean pooling. The results shown in \cref{tab:pooling} illustrate that the method based on concatenation achieves better performance.

\begin{table}[htbp]
  \centering
    \begin{tabular}{c|c|c}
    \hline
    Method & Paragraph feature & CSV \\
    \hline
    \multirow{2}[2]{*}{Ours} & pooling &67.07 \\
          & concat &67.46  \\
    \hline
    \end{tabular}%
      \caption{Ablation studies of the ways to extract paragraph features on our method.}
  \label{tab:pooling}%
\end{table}

\subsection{Sequence alignment loss}
For a fair comparison, some adjustments have been made to the architecture of our model on the supervised sequence verification task. Specifically, following \cite{SVIP}, we apply the video sequence alignment mechanism to our model. Moreover, we also conduct experiments to investigate the effectiveness of using sequence alignment loss. We change the sequence align loss position to the last of the network. The results shown in \cref{tab:seq_loss} illustrate that sequence alignment loss $L_{\text{seq}}$ could restrict the model to learning a  better representation. 
\begin{table}[htbp]
  \centering

    \begin{tabular}{c|c|c}
    \hline
    Method & $L_{\text{seq}}$ & CSV \\
    \hline
    \multirow{2}[2]{*}{Ours} & \graycross &84.47  \\
          & \ding{51} &84.69  \\
    \hline
    \end{tabular}%
      \caption{Ablation studies of the sequence alignment loss on our method.}
  \label{tab:seq_loss}%
\end{table}%

\section{Gumbel-Softmax with Viterbi}
Due to the sum of the probabilities of each row cannot be greater than one and each probability value in a row should be the same, we simply set the value to $\frac{1}{N}$. 
As \cref{eq:viterbi_transition2} shown, we set each element value in the upper diagonal matrix to $\frac{1}{N}$ and others to zero to keep the path of probability will be a one-way path.
\begin{equation}
    A=
    \begin{bmatrix}
        \frac{1}{N} & \dots  & \frac{1}{N} \\
               & \ddots & \vdots \\
        0      &        & \frac{1}{N}\\
    \end{bmatrix}_{N \times N}
    \label{eq:viterbi_transition2}
\end{equation}
where $A$ represents the Transition matrix of Viterbi algorithm\cite{viterbi}.

\section{TSM module}
Following \cite{RepNet}, we add the Temporal Similarity Matrix (TSM) module with residual connection to our vision module. In this ablation study, we only use the task classification loss $L_\text{cls}$ instead of coarse-grained loss $L_\text{coarse}$ and fine-grained loss $L_\text{fine}$. As \cref{tab:tsm_type} shown, we verify different similarity distances of TSM and residual connection types. And the experiments indicate that the TSM module with residual connection will improve the model performance. 

However, as \cref{tab:add_tsm} shows, while we apply the TSM module to our method and train the model under weak supervision, the performance of the model degrades. 
It is reasonable that the model with the TSM module is not effective for language-video alignment tasks.

\begin{table}[htbp]
  \centering

    \begin{tabular}{c|cc|c}
    \hline
    Method &  Dist & Residual & CSV \\
    \hline
    \multirow{5}[2]{*}{CLIP\cite{CLIP}+TE\cite{vit}+MLP} 
            & \graycross &\graycross &77.35  \\
    \cline{2-4}
            & L2    & add   & 77.42 \\
          & L2    & concat & 78.22 \\
          & Attn  & add   & 76.89 \\
          & Attn  & concat & 77.71 \\
    \hline
    \end{tabular}%
      \caption{Ablation studies of the different kinds of TSM module on the baseline.}
  \label{tab:tsm_type}%
\end{table}%

\begin{table}[htbp]
  \centering

    \begin{tabular}{c|c|c|c|c}
    \hline
    Method & $L_\text{fine}$ & $L_\text{coarse}$ & TSM   & CSV \\
    \hline
    \multirow{2}[4]{*}{Ours} & \ding{51} & \ding{51} & \graycross & 79.80 \\
\cline{2-5}          & \ding{51} & \ding{51} & \ding{51} & 76.00  \\
    \hline
    \end{tabular}%
      \caption{Ablation studies of the TSM on our proposed method.}
  \label{tab:add_tsm}%
\end{table}%
\begin{table*}
\centering
\begin{tabular}{c|c|c|c}
\hline
Method                        & Backbone  & Loss                         & Classification(Acc)\\
\hline
CAT\cite{SVIP}                           & ResNet-50\cite{RepNet} & CLS, SEQ                     & 61.08 \\
CLIP\cite{CLIP}+TE+MLP                   & CLIP-ViT  & CLS, SEQ                     & 63.24 \\
\hline
Ours(CLS)                     & CLIP-ViT  & CLS, SEQ, Multi-grained loss & \textbf{69.57}       \\ 
\hline
\end{tabular}%
\caption{Results of video classification on CSV.}
\label{tab:classification}
\end{table*}

\begin{table*}[htbp]
\centering

\begin{tabular}{c|c|ccc|ccc}
	\hline
	\multirow{2}[4]{*}{Method}           & \multirow{2}[4]{*}{Backbone} & \multicolumn{3}{c|}{Weakly supervised (w/o CLS)} & \multicolumn{3}{c}{Supervised (w CLS)}                                                                                      \\
	\cline{3-8}                          &                              & Def.                                             & No Rep.                                          & Rep.            & Def.            & No Rep.         & Rep.               \\
	\hline
	CAT\cite{SVIP}                           & ResNet50\cite{ResNet}       & 47.70                                            & 57.82                                            & 49.99           & 51.13           & 63.25           & 45.96           \\
	CLIP\cite{CLIP}+TE+MLP & CLIP-ViT       & 50.83                                            & 65.28                                            & 53.73           & 48.50           & 65.21           & 51.25                        \\
    \hline
	Ours                                 & CLIP-ViT       & \textbf{52.55 }                                  & \textbf{68.98 }                                  & \textbf{56.16 } & \textbf{59.57 } & \textbf{77.78 } & \textbf{54.95 }  \\
	\hline
\end{tabular}%
\caption{Results of different methods on re-divided COIN-SV.}
  \label{tab:limitation}%
\end{table*}%
\section{Downstream tasks}
\subsection{Text-to-Video Matching}

We validate the performance of the video-language representations on text-to-video matching, which aims to find the correct video corresponding to a sequence of texts from a series of videos. Specifically, we train our model on the CSV dataset under weak supervision and test it on our proposed benchmark about text-to-video matching. We calculate the similarity between each video representation $V_i$ and paragraph representation $L$:
\begin{equation}
    d = dis(L,V_i)  \\
\end{equation}
where $dis(.,.)$ represents the normalized Euclidean distance. And $V_i$ represents $i_{th}(i \in [0,\dots,4])$ video representation.
At last, we select the text-video pair with the max similarity. 

\noindent \textbf{CSV-Matching}. 
To better evaluate the text-to-video matching, we rearrange the test set of CSV and propose a new scripted benchmark named CSV-Matching. 
It has 800 text-video pairs. Each text-video pair is composed of one sequence of text descriptions of procedures and five videos. All of the videos describe the same task but hold different procedures. There is only one correct video matching the text descriptions in each pair. CSV-test dataset contains 5 tasks and each task has 5 kinds of different procedures. We random select one kind of video from each procedure to compose one pairs. The benchmark and split script will be available.

\subsection{Video Classification}

To demonstrate our method's transfer ability, we evaluate models in the downstream video classification task. We re-divided the CSV dataset for video classification task. The train set contains 689 videos and test set contains 185 videos.
On the re-divided CSV test dataset (CSV-CLS), we evaluate representations of models with linear probing, which were pre-trained under weak supervision. As \cref{tab:classification} shown, our method achieves better performance in the video classification task. The benchmark and split script will be available.

\section{Limitations}

While our method performs well on the major part of the data, there still are some failure cases. In realistic sequential videos, sub-actions are often repeated. In that case, there are multiple sentences with high similarity to a frame. It could mislead the model to generate biased pseudo-labels, which will lead to the deterioration of performance. For example, the occurrence of a large number of repetitive actions repetitive action might hidden achieving further performance. 

The intuition of our fine-grained contrastive loss comes from a basic idea: \textit{if the $s_j$ is the corresponding sentence for frame $ h_i$, the corresponding sentence for frame $h_{i+1}$ is never before the $s_j$ in sequence}. Due to a large number of repetitive actions, it might be difficult to achieve further performance. However, this method is still promising. As \cref{tab:limitation} shown, we have re-divided the COIN-SV test dataset based on whether existing repetitive actions in videos or not, which are COIN-SV-Rep (675 video pairs) and COIN-SV-NoRep (325 video pairs). In the original COIN-SV test dataset, there are 1000 video pairs for sequential video verification, built by 328 videos containing repetitive actions and 123 videos that do not. The results show that although the occurrence of repetitive actions will cause the deterioration of performance, our method can still achieve better results than other baselines. Moreover, the results conducted by our method may reflect the bias from the dataset. 

{\small
\bibliographystyle{ieee_fullname}
\bibliography{References}

\begin{thebibliography}{10}\itemsep=-1pt

\bibitem{Timestamp}
Nadine Behrmann, S~Alireza Golestaneh, Zico Kolter, Juergen Gall, and Mehdi
  Noroozi.
\newblock Unified fully and timestamp supervised temporal action segmentation
  via sequence to sequence translation.
\newblock In {\em European Conference on Computer Vision}, pages 52--68.
  Springer, 2022.

\bibitem{ikea}
Yizhak Ben-Shabat, Xin Yu, Fatemeh Saleh, Dylan Campbell, Cristian
  Rodriguez-Opazo, Hongdong Li, and Stephen Gould.
\newblock The ikea asm dataset: Understanding people assembling furniture
  through actions, objects and pose.
\newblock In {\em Proceedings of the IEEE/CVF Winter Conference on Applications
  of Computer Vision}, pages 847--859, 2021.

\bibitem{timesformer}
Gedas Bertasius, Heng Wang, and Lorenzo Torresani.
\newblock Is space-time attention all you need for video understanding?
\newblock In {\em ICML}, volume~2, page~4, 2021.

\bibitem{LocalVTP}
Meng Cao, Tianyu Yang, Junwu Weng, Can Zhang, Jue Wang, and Yuexian Zou.
\newblock Locvtp: Video-text pre-training for temporal localization.
\newblock {\em arXiv preprint arXiv:2207.10362}, 2022.

\bibitem{kinetics}
Joao Carreira and Andrew Zisserman.
\newblock Quo vadis, action recognition? a new model and the kinetics dataset.
\newblock In {\em proceedings of the IEEE Conference on Computer Vision and
  Pattern Recognition}, pages 6299--6308, 2017.

\bibitem{chang2019d3tw}
Chien-Yi Chang, De-An Huang, Yanan Sui, Li Fei-Fei, and Juan~Carlos Niebles.
\newblock D3tw: Discriminative differentiable dynamic time warping for weakly
  supervised action alignment and segmentation.
\newblock In {\em Proceedings of the IEEE/CVF Conference on Computer Vision and
  Pattern Recognition}, pages 3546--3555, 2019.

\bibitem{frame-sequence}
Minghao Chen, Fangyun Wei, Chong Li, and Deng Cai.
\newblock Frame-wise action representations for long videos via sequence
  contrastive learning.
\newblock In {\em Proceedings of the IEEE/CVF Conference on Computer Vision and
  Pattern Recognition}, pages 13801--13810, 2022.

\bibitem{SimCLR}
Ting Chen, Simon Kornblith, Mohammad Norouzi, and Geoffrey Hinton.
\newblock A simple framework for contrastive learning of visual
  representations.
\newblock In {\em International conference on machine learning}, pages
  1597--1607. PMLR, 2020.

\bibitem{epic-kitchens}
Dima Damen, Hazel Doughty, Giovanni~Maria Farinella, Sanja Fidler, Antonino
  Furnari, Evangelos Kazakos, Davide Moltisanti, Jonathan Munro, Toby Perrett,
  Will Price, et~al.
\newblock Scaling egocentric vision: The epic-kitchens dataset.
\newblock In {\em Proceedings of the European Conference on Computer Vision
  (ECCV)}, pages 720--736, 2018.

\bibitem{arcface}
Jiankang Deng, Jia Guo, Niannan Xue, and Stefanos Zafeiriou.
\newblock Arcface: Additive angular margin loss for deep face recognition.
\newblock In {\em Proceedings of the IEEE/CVF Conference on Computer Vision and
  Pattern Recognition (CVPR)}, June 2019.

\bibitem{vit}
Alexey Dosovitskiy, Lucas Beyer, Alexander Kolesnikov, Dirk Weissenborn,
  Xiaohua Zhai, Thomas Unterthiner, Mostafa Dehghani, Matthias Minderer, Georg
  Heigold, Sylvain Gelly, et~al.
\newblock An image is worth 16x16 words: Transformers for image recognition at
  scale.
\newblock {\em arXiv preprint arXiv:2010.11929}, 2020.

\bibitem{RepNet}
Debidatta Dwibedi, Yusuf Aytar, Jonathan Tompson, Pierre Sermanet, and Andrew
  Zisserman.
\newblock Counting out time: Class agnostic video repetition counting in the
  wild.
\newblock In {\em Proceedings of the IEEE/CVF conference on computer vision and
  pattern recognition}, pages 10387--10396, 2020.

\bibitem{MViT}
Haoqi Fan, Bo Xiong, Karttikeya Mangalam, Yanghao Li, Zhicheng Yan, Jitendra
  Malik, and Christoph Feichtenhofer.
\newblock Multiscale vision transformers.
\newblock In {\em Proceedings of the IEEE/CVF International Conference on
  Computer Vision}, pages 6824--6835, 2021.

\bibitem{slowfast}
Christoph Feichtenhofer, Haoqi Fan, Jitendra Malik, and Kaiming He.
\newblock Slowfast networks for video recognition.
\newblock In {\em Proceedings of the IEEE/CVF international conference on
  computer vision}, pages 6202--6211, 2019.

\bibitem{mist}
Jia-Chang Feng, Fa-Ting Hong, and Wei-Shi Zheng.
\newblock Mist: Multiple instance self-training framework for video anomaly
  detection.
\newblock In {\em Proceedings of the IEEE/CVF conference on computer vision and
  pattern recognition}, pages 14009--14018, 2021.

\bibitem{viterbi}
G~David Forney.
\newblock The viterbi algorithm.
\newblock {\em Proceedings of the IEEE}, 61(3):268--278, 1973.

\bibitem{bridgeformer}
Yuying Ge, Yixiao Ge, Xihui Liu, Dian Li, Ying Shan, Xiaohu Qie, and Ping Luo.
\newblock Bridgeformer: Bridging video-text retrieval with multiple choice
  questions.
\newblock {\em arXiv preprint arXiv:2201.04850}, 2022.

\bibitem{Xie2022Alignment}
Tengda Han, Weidi Xie, and Andrew Zisserman.
\newblock Temporal alignment networks for long-term video.
\newblock In {\em Proceedings of the IEEE/CVF Conference on Computer Vision and
  Pattern Recognition}, pages 2906--2916, 2022.

\bibitem{MAE}
Kaiming He, Xinlei Chen, Saining Xie, Yanghao Li, Piotr Doll{\'a}r, and Ross
  Girshick.
\newblock Masked autoencoders are scalable vision learners.
\newblock In {\em Proceedings of the IEEE/CVF Conference on Computer Vision and
  Pattern Recognition}, pages 16000--16009, 2022.

\bibitem{kaiming}
Kaiming He, Xiangyu Zhang, Shaoqing Ren, and Jian Sun.
\newblock Delving deep into rectifiers: Surpassing human-level performance on
  imagenet classification.
\newblock In {\em Proceedings of the IEEE international conference on computer
  vision}, pages 1026--1034, 2015.

\bibitem{ResNet}
Kaiming He, Xiangyu Zhang, Shaoqing Ren, and Jian Sun.
\newblock Deep residual learning for image recognition.
\newblock In {\em Proceedings of the IEEE conference on computer vision and
  pattern recognition}, pages 770--778, 2016.

\bibitem{hu2022transrac}
Huazhang Hu, Sixun Dong, Yiqun Zhao, Dongze Lian, Zhengxin Li, and Shenghua
  Gao.
\newblock Transrac: Encoding multi-scale temporal correlation with transformers
  for repetitive action counting.
\newblock In {\em Proceedings of the IEEE/CVF Conference on Computer Vision and
  Pattern Recognition}, pages 19013--19022, 2022.

\bibitem{gumbel_softmax}
Eric Jang, Shixiang Gu, and Ben Poole.
\newblock Categorical reparameterization with gumbel-softmax.
\newblock {\em arXiv preprint arXiv:1611.01144}, 2016.

\bibitem{review1_1}
Anna Kukleva, Hilde Kuehne, Fadime Sener, and Jurgen Gall.
\newblock Unsupervised learning of action classes with continuous temporal
  embedding.
\newblock In {\em Proceedings of the IEEE/CVF Conference on Computer Vision and
  Pattern Recognition}, pages 12066--12074, 2019.

\bibitem{kumar2022unsupervised}
Sateesh Kumar, Sanjay Haresh, Awais Ahmed, Andrey Konin, M~Zeeshan Zia, and
  Quoc-Huy Tran.
\newblock Unsupervised action segmentation by joint representation learning and
  online clustering.
\newblock In {\em Proceedings of the IEEE/CVF Conference on Computer Vision and
  Pattern Recognition}, pages 20174--20185, 2022.

\bibitem{segmentation}
Colin Lea, Michael~D Flynn, Rene Vidal, Austin Reiter, and Gregory~D Hager.
\newblock Temporal convolutional networks for action segmentation and
  detection.
\newblock In {\em proceedings of the IEEE Conference on Computer Vision and
  Pattern Recognition}, pages 156--165, 2017.

\bibitem{review1_3}
Jun Li and Sinisa Todorovic.
\newblock Action shuffle alternating learning for unsupervised action
  segmentation.
\newblock In {\em Proceedings of the IEEE/CVF Conference on Computer Vision and
  Pattern Recognition}, pages 12628--12636, 2021.

\bibitem{clip-event}
Manling Li, Ruochen Xu, Shuohang Wang, Luowei Zhou, Xudong Lin, Chenguang Zhu,
  Michael Zeng, Heng Ji, and Shih-Fu Chang.
\newblock Clip-event: Connecting text and images with event structures.
\newblock In {\em Proceedings of the IEEE/CVF Conference on Computer Vision and
  Pattern Recognition}, pages 16420--16429, 2022.

\bibitem{Diving}
Yingwei Li, Yi Li, and Nuno Vasconcelos.
\newblock Resound: Towards action recognition without representation bias.
\newblock In {\em Proceedings of the European Conference on Computer Vision
  (ECCV)}, pages 513--528, 2018.

\bibitem{cross-modal-rl}
Chung-Ching Lin, Kevin Lin, Lijuan Wang, Zicheng Liu, and Linjie Li.
\newblock Cross-modal representation learning for zero-shot action recognition.
\newblock In {\em Proceedings of the IEEE/CVF Conference on Computer Vision and
  Pattern Recognition}, pages 19978--19988, 2022.

\bibitem{procedureactivity}
Xudong Lin, Fabio Petroni, Gedas Bertasius, Marcus Rohrbach, Shih-Fu Chang, and
  Lorenzo Torresani.
\newblock Learning to recognize procedural activities with distant supervision.
\newblock In {\em Proceedings of the IEEE/CVF Conference on Computer Vision and
  Pattern Recognition}, pages 13853--13863, 2022.

\bibitem{swin}
Ze Liu, Yutong Lin, Yue Cao, Han Hu, Yixuan Wei, Zheng Zhang, Stephen Lin, and
  Baining Guo.
\newblock Swin transformer: Hierarchical vision transformer using shifted
  windows.
\newblock In {\em Proceedings of the IEEE/CVF International Conference on
  Computer Vision}, pages 10012--10022, 2021.

\bibitem{video-swin}
Ze Liu, Jia Ning, Yue Cao, Yixuan Wei, Zheng Zhang, Stephen Lin, and Han Hu.
\newblock Video swin transformer.
\newblock In {\em Proceedings of the IEEE/CVF Conference on Computer Vision and
  Pattern Recognition}, pages 3202--3211, 2022.

\bibitem{set_supervised}
Zijia Lu and Ehsan Elhamifar.
\newblock Set-supervised action learning in procedural task videos via pairwise
  order consistency.
\newblock In {\em Proceedings of the IEEE/CVF Conference on Computer Vision and
  Pattern Recognition}, pages 19903--19913, 2022.

\bibitem{clip4clip}
Huaishao Luo, Lei Ji, Ming Zhong, Yang Chen, Wen Lei, Nan Duan, and Tianrui Li.
\newblock Clip4clip: An empirical study of clip for end to end video clip
  retrieval and captioning.
\newblock {\em Neurocomputing}, 508:293--304, 2022.

\bibitem{MIL-NCE}
Antoine Miech, Jean-Baptiste Alayrac, Lucas Smaira, Ivan Laptev, Josef Sivic,
  and Andrew Zisserman.
\newblock End-to-end learning of visual representations from uncurated
  instructional videos.
\newblock In {\em Proceedings of the IEEE/CVF Conference on Computer Vision and
  Pattern Recognition}, pages 9879--9889, 2020.

\bibitem{howto100m}
Antoine Miech, Dimitri Zhukov, Jean-Baptiste Alayrac, Makarand Tapaswi, Ivan
  Laptev, and Josef Sivic.
\newblock Howto100m: Learning a text-video embedding by watching hundred
  million narrated video clips.
\newblock In {\em Proceedings of the IEEE/CVF International Conference on
  Computer Vision}, pages 2630--2640, 2019.

\bibitem{mintz2009distant}
Mike Mintz, Steven Bills, Rion Snow, and Dan Jurafsky.
\newblock Distant supervision for relation extraction without labeled data.
\newblock In {\em Proceedings of the Joint Conference of the 47th Annual
  Meeting of the ACL and the 4th International Joint Conference on Natural
  Language Processing of the AFNLP}, pages 1003--1011, 2009.

\bibitem{X-CLIP}
Bolin Ni, Houwen Peng, Minghao Chen, Songyang Zhang, Gaofeng Meng, Jianlong Fu,
  Shiming Xiang, and Haibin Ling.
\newblock Expanding language-image pretrained models for general video
  recognition.
\newblock In {\em European Conference on Computer Vision}, pages 1--18.
  Springer, 2022.

\bibitem{InfoNCE}
Aaron van~den Oord, Yazhe Li, and Oriol Vinyals.
\newblock Representation learning with contrastive predictive coding.
\newblock {\em arXiv preprint arXiv:1807.03748}, 2018.

\bibitem{SVIP}
Yicheng Qian, Weixin Luo, Dongze Lian, Xu Tang, Peilin Zhao, and Shenghua Gao.
\newblock Svip: Sequence verification for procedures in videos.
\newblock In {\em Proceedings of the IEEE/CVF Conference on Computer Vision and
  Pattern Recognition}, pages 19890--19902, 2022.

\bibitem{detection}
Fang Qingyun, Han Dapeng, and Wang Zhaokui.
\newblock Cross-modality fusion transformer for multispectral object detection.
\newblock {\em arXiv preprint arXiv:2111.00273}, 2021.

\bibitem{CLIP}
Alec Radford, Jong~Wook Kim, Chris Hallacy, Aditya Ramesh, Gabriel Goh,
  Sandhini Agarwal, Girish Sastry, Amanda Askell, Pamela Mishkin, Jack Clark,
  et~al.
\newblock Learning transferable visual models from natural language
  supervision.
\newblock In {\em International Conference on Machine Learning}, pages
  8748--8763. PMLR, 2021.

\bibitem{assembly101}
Fadime Sener, Dibyadip Chatterjee, Daniel Shelepov, Kun He, Dipika Singhania,
  Robert Wang, and Angela Yao.
\newblock Assembly101: A large-scale multi-view video dataset for understanding
  procedural activities.
\newblock In {\em Proceedings of the IEEE/CVF Conference on Computer Vision and
  Pattern Recognition}, pages 21096--21106, 2022.

\bibitem{MPNet}
Kaitao Song, Xu Tan, Tao Qin, Jianfeng Lu, and Tie-Yan Liu.
\newblock Mpnet: Masked and permuted pre-training for language understanding.
\newblock {\em Advances in Neural Information Processing Systems},
  33:16857--16867, 2020.

\bibitem{sun2019learning}
Chen Sun, Fabien Baradel, Kevin Murphy, and Cordelia Schmid.
\newblock Learning video representations using contrastive bidirectional
  transformer.
\newblock {\em arXiv preprint arXiv:1906.05743}, 2019.

\bibitem{videobert}
Chen Sun, Austin Myers, Carl Vondrick, Kevin Murphy, and Cordelia Schmid.
\newblock Videobert: A joint model for video and language representation
  learning.
\newblock In {\em Proceedings of the IEEE/CVF International Conference on
  Computer Vision}, pages 7464--7473, 2019.

\bibitem{sun2022long}
Yuchong Sun, Hongwei Xue, Ruihua Song, Bei Liu, Huan Yang, and Jianlong Fu.
\newblock Long-form video-language pre-training with multimodal temporal
  contrastive learning.
\newblock {\em arXiv preprint arXiv:2210.06031}, 2022.

\bibitem{coin}
Yansong Tang, Dajun Ding, Yongming Rao, Yu Zheng, Danyang Zhang, Lili Zhao,
  Jiwen Lu, and Jie Zhou.
\newblock Coin: A large-scale dataset for comprehensive instructional video
  analysis.
\newblock In {\em Proceedings of the IEEE/CVF Conference on Computer Vision and
  Pattern Recognition}, pages 1207--1216, 2019.

\bibitem{C3D}
Du Tran, Lubomir Bourdev, Rob Fergus, Lorenzo Torresani, and Manohar Paluri.
\newblock Learning spatiotemporal features with 3d convolutional networks.
\newblock In {\em Proceedings of the IEEE international conference on computer
  vision}, pages 4489--4497, 2015.

\bibitem{tran2018closer}
Du Tran, Heng Wang, Lorenzo Torresani, Jamie Ray, Yann LeCun, and Manohar
  Paluri.
\newblock A closer look at spatiotemporal convolutions for action recognition.
\newblock In {\em Proceedings of the IEEE conference on Computer Vision and
  Pattern Recognition}, pages 6450--6459, 2018.

\bibitem{transformer}
Ashish Vaswani, Noam Shazeer, Niki Parmar, Jakob Uszkoreit, Llion Jones,
  Aidan~N Gomez, {\L}ukasz Kaiser, and Illia Polosukhin.
\newblock Attention is all you need.
\newblock {\em Advances in neural information processing systems}, 30, 2017.

\bibitem{wang2022long}
Jue Wang, Gedas Bertasius, Du Tran, and Lorenzo Torresani.
\newblock Long-short temporal contrastive learning of video transformers.
\newblock In {\em Proceedings of the IEEE/CVF Conference on Computer Vision and
  Pattern Recognition}, pages 14010--14020, 2022.

\bibitem{localization}
Limin Wang, Yuanjun Xiong, Zhe Wang, Yu Qiao, Dahua Lin, Xiaoou Tang, and Luc
  Van~Gool.
\newblock Temporal segment networks: Towards good practices for deep action
  recognition.
\newblock In {\em European conference on computer vision}, pages 20--36.
  Springer, 2016.

\bibitem{TSN}
Limin Wang, Yuanjun Xiong, Zhe Wang, Yu Qiao, Dahua Lin, Xiaoou Tang, and Luc
  Van~Gool.
\newblock Temporal segment networks: Towards good practices for deep action
  recognition.
\newblock In {\em European conference on computer vision}, pages 20--36.
  Springer, 2016.

\bibitem{classification}
Limin Wang, Yuanjun Xiong, Zhe Wang, Yu Qiao, Dahua Lin, Xiaoou Tang, and Luc
  Van~Gool.
\newblock Temporal segment networks for action recognition in videos.
\newblock {\em IEEE transactions on pattern analysis and machine intelligence},
  41(11):2740--2755, 2018.

\bibitem{actionclip}
Mengmeng Wang, Jiazheng Xing, and Yong Liu.
\newblock Actionclip: A new paradigm for video action recognition.
\newblock {\em arXiv preprint arXiv:2109.08472}, 2021.

\bibitem{VLM}
Hu Xu, Gargi Ghosh, Po-Yao Huang, Prahal Arora, Masoumeh Aminzadeh, Christoph
  Feichtenhofer, Florian Metze, and Luke Zettlemoyer.
\newblock Vlm: Task-agnostic video-language model pre-training for video
  understanding.
\newblock {\em arXiv preprint arXiv:2105.09996}, 2021.

\bibitem{videoclip}
Hu Xu, Gargi Ghosh, Po-Yao Huang, Dmytro Okhonko, Armen Aghajanyan, Florian
  Metze, Luke Zettlemoyer, and Christoph Feichtenhofer.
\newblock Videoclip: Contrastive pre-training for zero-shot video-text
  understanding.
\newblock {\em arXiv preprint arXiv:2109.14084}, 2021.

\bibitem{adamw}
Yang You, Jing Li, Sashank Reddi, Jonathan Hseu, Sanjiv Kumar, Srinadh
  Bhojanapalli, Xiaodan Song, James Demmel, Kurt Keutzer, and Cho-Jui Hsieh.
\newblock Large batch optimization for deep learning: Training bert in 76
  minutes.
\newblock {\em arXiv preprint arXiv:1904.00962}, 2019.

\bibitem{coca}
Jiahui Yu, Zirui Wang, Vijay Vasudevan, Legg Yeung, Mojtaba Seyedhosseini, and
  Yonghui Wu.
\newblock Coca: Contrastive captioners are image-text foundation models.
\newblock {\em arXiv preprint arXiv:2205.01917}, 2022.

\bibitem{zeng2015distant}
Daojian Zeng, Kang Liu, Yubo Chen, and Jun Zhao.
\newblock Distant supervision for relation extraction via piecewise
  convolutional neural networks.
\newblock In {\em Proceedings of the 2015 conference on empirical methods in
  natural language processing}, pages 1753--1762, 2015.

\bibitem{CAT-Det}
Yanan Zhang, Jiaxin Chen, and Di Huang.
\newblock Cat-det: Contrastively augmented transformer for multi-modal 3d
  object detection.
\newblock In {\em Proceedings of the IEEE/CVF Conference on Computer Vision and
  Pattern Recognition}, pages 908--917, 2022.

\bibitem{TRN}
Bolei Zhou, Alex Andonian, Aude Oliva, and Antonio Torralba.
\newblock Temporal relational reasoning in videos.
\newblock In {\em Proceedings of the European conference on computer vision
  (ECCV)}, pages 803--818, 2018.

\bibitem{actbert}
Linchao Zhu and Yi Yang.
\newblock Actbert: Learning global-local video-text representations.
\newblock In {\em Proceedings of the IEEE/CVF conference on computer vision and
  pattern recognition}, pages 8746--8755, 2020.

\end{thebibliography}
}

\end{document}